\documentclass[conference]{IEEEtran}

\IEEEoverridecommandlockouts

\usepackage{amsmath,amssymb,amsfonts}
\usepackage{graphicx,textcomp}
\usepackage{subfig,float}
\usepackage{algorithmic,algorithm}
\usepackage{multirow,array}
\usepackage{hyperref,cite,url}
\usepackage{color}
\usepackage{comment}

\newcommand{\bx}{\mathbf{x}}

\newcommand{\bz}{\mathbf{z}}

\newcommand{\bn}{\mathbf{n}}
\newcommand{\bv}{\mathbf{v}}

\newcommand{\bzt}{\bz^{(\text{t})}}
\newcommand{\bzc}{\bz^{(\text{c})}}
\newcommand{\bzzt}{\mathbf{Z}^{(\text{t})}}
\newcommand{\bzzc}{\mathbf{Z}^{(\text{c})}}

\newcommand{\bmu}{\boldsymbol{\mu}}

\newcommand{\eqq}{$\,=\,$}
\newcommand{\no}{\noindent}

\newcommand{\ben}{\begin{eqnarray*}}
\newcommand{\een}{\end{eqnarray*}}

\newcommand{\be}{\begin{eqnarray}}
\newcommand{\ee}{\end{eqnarray}}

\usepackage{bbm}

\newcommand{\prob}{\mathbb{P}}

 \IEEEoverridecommandlockouts

\begin{document}

\title{\LARGE  Use of Bayesian Nonparametric methods for Estimating the Measurements in High Clutter.   }

\author{Bahman Moraffah$^\dag$, Christ Richmond$^\dag$, Raha Moraffah$^\ddag$, and Antonia Papandreou-Suppappola$^\dag$ \\
\IEEEauthorblockA{$^\dag$Department of Electrical, Computer, and Energy, 
Arizona State University, Tempe AZ\thanks{This work was 
supported in part by Grant AFOSR FA9550-17-1-0100.}\\
$^\ddag$ School of Computing, Informatics, and Decision Systems Engineering, Arizona State University, Tempe AZ\\
\{bahman.moraffah, christ.richmond, raha.moraffah, papandreou\}@asu.edu}}

\maketitle


\begin{abstract}
  
Robust tracking of a target in a clutter environment is an important and challenging task. In recent years, the nearest neighbor methods and probabilistic data association filters were proposed. However, the performance of these methods diminishes as number of measurements increases. In this paper, we propose a robust generative approach to effectively model multiple sensor measurements for tracking a moving target in an environment with high clutter. We assume a time-dependent number of measurements that include sensor observations with unknown origin, some of which may only contain clutter with no additional information. We robustly and accurately estimate the trajectory of the moving target in high clutter environment with unknown number of clutters by employing Bayesian nonparametric modeling. In particular,  we employ a class of joint Bayesian nonparametric models to construct the joint prior distribution of target and clutter measurements such that the conditional distributions follow a Dirichlet process. The marginalized Dirichlet process prior of the target measurements is then used in a Bayesian tracker to estimate the dynamically-varying target state. We show through experiments that the tracking performance and effectiveness of our proposed framework are increased by suppressing high clutter measurements. In addition, we show that our proposed method outperforms existing methods such as nearest neighbor and probability data association filters.

\end{abstract}


\section{Introduction}
Target tracking in a high clutter environment 
  is very challenging as sensor measurements , more often than not, contain detections from false targets \cite{bar2011tracking}. Time-dependent number of measurements that include both clutter and true sensor observations with unknown origin makes the tracking in the clutter environment difficult. To solve this problem a set of assumptions is made: 1) true measurements from the target are present with some probability of detection, and 2) number and location of clutter measurements are unknown and random.  

  For accurate target parameter estimates,  tracking algorithms must only 
 incorporate target generated measurements. To this end, various methods have been considered to solve the problem of tracking in clutter. One of the widely used methods is the strongest-neighbor and nearest-neighbor (NN) filters \cite{davis1978noise}. In this method, a Gaussian object motion considered and measurements that are statistically closest to the predicted measurements are from the object and the rest are considered clutters. It is shown that these filters
only process one measurement,  the one with the highest probability that is generated by the target \cite{Li98}; however,  the 
 performance of these methods diminish with increase in the probability of false alarm rate. 
 
  In \cite{Kir04,bar2011tracking},  the probabilistic data association (PDA) method 
  was used  to validate multiple measurements according to their probability 
  of target origin. Similar to nearest-neighbor filters, PDA filters assume that the object motion obeys linear Gaussian statistics. The PDA filter assumes that all non-object originated measurements are clutter which is uniformly distributed in the space and Poisson distributed in time. Furthermore, several variations of the PDA method are proposed, such as 
  the filtered gate structure method, the interactive-multiple model PDA 
  \cite{Jeo05}, as well as other data association approaches,  
  such as the Viterbi  and  fuzzy data association,  
  were also considered \cite{gad2002comparison}.
  Note, however, that PDA type methods can become 
computationally intensive as the number of measurements increases.

 Recently,  Bayesian nonparametric methods \cite{hjort2010bayesian, moraffah2019bayesian}
  have  been used in target tracking to solve problems 
 such as multiple models \cite{caron2007bayesian}, maneuvering \cite{Fox11}, dependent measurements \cite{Mor19b, moraffah2020bayesian, moraffah2019nonparametric}, waveform design \cite{wang18}, and 
measurement-to-target associations in multiple target tracking \cite{Mor18,Mor19a, moraffah2019inference}.
 In this paper,  we propose the measurement estimation for tracking in high clutter (METRIC) Bayesian nonparametric approach to track a target in high clutter.  Our proposed model does not suffer from the issues that aforementioned methods face and is computationally inexpensive. 
In particular,  the joint prior distribution 
 of measurements with different origins is modeled
 using Dirichlet processes (DPs).
 The marginal DP prior of the target measurements 
 is then integrated into Bayesian filtering methods to track the target. 
 
 The rest of the paper is organized as follows. In Section \ref{section_review}, we briefly review Bayesian nonprahmteric methods. In particular, we study the Dirichlet process which is used in the rest of this paper. Section \ref{section_model} discusses the joint modeling of the measurements through a joint Dirichlet process. We integrate the proposed model into a Bayesian tracker to robustly and accurately track the target trajectory. Section \ref{sim} demonstrates the advantage of the proposed model over the existing methods.

\section{Background: Bayesian Nonparametric Modeling}
\label{section_review}

\no With the surge of Monte Carlo (MC) methods, Bayesian nonparametric models have been ubiquitous in probabilistic modeling of the observations to avoid the limitations in parametric modeling such as model misfit. The family of infinite dimension of random measures has been used as a substitute of finite mixture model estimate the density. In this section, we briefly describe one of the main stochastic processes that play an integral role in our investigation. 

\subsection{Dirichlet Process}
Dirichlet process (DP) is the most well-studied nonparametric model that defines a prior over the space of probability distributions \cite{ferguson1973bayesian}. Dirichlet process can be considered as the infinite generalization of the Dirichlet distribution, that is 
\begin{equation}
\label{dp1}
\lim\limits_{K\to\infty} \sum\limits_{k=1}^{K}  \pi_k \delta_{\theta_k}
\end{equation}
 for mixing distribution $\pi = (\pi_1, \dots, \pi_K)$ and $\delta_\theta$ is a point mass centered at the parameter $\theta \in \Theta$. It can be shown equation (\ref{dp1}) results in a well-defined Dirichlet process \cite{sethuraman1994constructive, ghosh2003bayesian}. Dirichlet process is discrete with probability one and thus is mainly used as a prior on the parameters in the space of parameters $\Theta$. A Dirichlet process with concentration parameter $\alpha$ and base distribution $H$ is denoted by $\text{DP}(\alpha, H)$. A draw $ G\sim \text{DP}(\alpha, H)$ is a probability random measure and is represented as  \cite{sethuraman1994constructive}
 \begin {equation}
 \label{stick}
 G(\cdot) = \sum\limits_{k=1}^{\infty}  \pi_k \delta_{\theta_k}(\cdot)
\end{equation}
for mixing probability $\pi \sim \text{GEM}(\alpha)$ and the sequence of $\theta_k$ is drawn independently and identically from H, i.e., $\theta_k\sim H$. Note that since $G$ provides a distribution on $\Theta$, $\theta_k$ takes on values in $\Theta$. The GEM($\alpha$) is defined as 
\begin{flalign}
\pi \sim& \text{ GEM}(\alpha)\Longleftrightarrow\\
 v_k \sim \text{Beta}(1, \alpha),& \hspace{0.1cm}\pi_k = v_k\prod\limits_{j<k}(1-v_k)\hspace{0.3cm} \text{for  }\hspace{0.1cm}{k = 1, 2, \dots}\notag.
\end{flalign}
This construction is known as stick-breaking construction of a Dirichlet process. Let $\Xi = \{\theta_1, \dots, \theta_N\}$ be the independent draws from $G$. 

\subsection{Inference}
The posterior distribution of $G$ given the observations $\Xi = \{\theta_1,\dots, \theta_N\}$ is a Dirichlet process with concentration parameter $\alpha_{\text{post}}$ and base probability measure $H_{\text{post}}$ , 
\begin{equation}
G | \Xi \sim \text{DP}(\alpha_{\text{post}}, H_{\text{post}})
\end{equation}
where $\alpha_{\text{post}} = \alpha + N$ and 
\begin{equation}
\label{dppost}
H_{\text{post}} = \frac{N}{\alpha+N}\frac{\sum_{k=1}^N\delta_{\theta_k}}{N} +  \frac{\alpha}{\alpha+N} H.
\end{equation}
where $\delta_{\theta}$ is a degenerate measure concentrated at $\theta$.

Definition of Dirichlet process implies that the posterior mean is $H_{\text{post}}$ which is a convex combination of prior distribution and the empirical distribution. Note that if $N\gg \alpha$ the second term in (\ref{dppost}) vanishes and the posterior base measure converges to the empirical distribution and therefore dominates the posterior distribution. This fact implies the consistency of the posterior distribution. 

The predictive distribution $\prob(\theta_{N+1}\in A | \Xi)$ for some measurable event $A\subset \Theta$ is computed with marginalizing out $G$ as
\be
\label{dppred}
\begin{split}
\prob(\theta_{N+1}\in A | \Xi) &= \mathbb{E}(\theta_{N+1} \mathbbm{1}_{\theta_{N+1}}(A) | \Xi) \\
&= \frac{\sum_{k=1}^N \delta_{\theta_k}(A) + \alpha H(A) } {\alpha+N}
\end{split}
\ee
where $\mathbbm{1}_{\theta_{N+1}}(A)$ is the indicator function, i.e., $\mathbbm{1}_{\theta_{N+1}}(A) = 1$ if $\theta_{N+1}\in A$ and $\mathbbm{1}_{\theta_{N+1}}(A) = 0$ if $\theta_{N+1}\notin A$. One can observe that drawing from the predictive distribution of $\theta_{N+1}$ given $\Xi$ is equivalent to draw $\theta_{N+1}\sim H_{\text{post}}$. Discreteness of the predictive distribution (\ref{dppred}) guarantees the repeat of draws. Let $n_k = \#\{\theta\in \Xi : \theta = \theta_k\}$, we can re-write (\ref{dppred}) as
\be
\label{dppred1}
\theta_{N+1} | \Xi \sim  \sum_{k=1}^K \frac{n_k} {\alpha+N}\delta_{\theta^*_k} + \frac{\alpha} {\alpha+N}H 
\ee
where $\Xi^ * = \{\theta^*_1, \dots, \theta^*_K\} \subset \Xi$ is the set of unique parameters in $\Xi$. Induced clustering distribution from distribution (\ref{dppred1}) is known as Chinese restaurant process (CRP). Intuitively speaking, a cluster with parameter $\theta_{N+1}$ is either identical to one of the elements of $\Xi^*$ with probability proportional to the popularity of each parameter $n_k$ or forms a new cluster, i.e., $\theta_{N+1}\sim H$ with probability proportional to $\alpha$. Moreover, it is shown that given $N$ points, the expected number of cluster is $\alpha \log(N)$ \cite{pitman2006combinatorial}.

\section{Bayesian Nonparametric Prior for Joint Measurements}
\label{section_model}
In Section \ref{problem_def}, we discuss the problem in hand. Our proposed solution through Bayesian nonparametric modeling is studied in Section \ref{problem_prior}. We introduce a joint Dirichlet process prior to model the measurements parameters as well the clutter. We, then in Section \ref{problem_tracker}, integrate our proposed method into a Bayesian tracker to efficiently estimate the target trajectory. 
\subsection{Object Tracking Modeling in Clutter}
\label{problem_def}
\no We consider the problem of tracking a target 
in an environment with high clutter, given multiple measurements. We aim to track a single target in clutter with unknown number of clutters.
We assume that the unknown target state parameter state vector  $\bx_k$ at time step $k$  is  given by 
$ \bx_k\eqq f(\bx_{k-1})+ \bn_k$, 
where $f(\bx_k)$ is the state transition function and  
 $\bn_k$ is the transition state error process. Note that the normality of the target motion is not required. 
The $n$th measurement vector $\bz_{n, k}$,
$n\eqq 1, \ldots,  N_k$,  is  
\be
\label{eq:measurement_model}
\bz_{n, k} =    \left \{ 
    \begin{array}{ll}
     \bzt_{n, k} = h(\bx_k) +\bv_{n, k},   & \text{Target  measurement}  \\[0.2in]
          \bzc_{n, k},  & \text{Clutter  measurement}
\end{array} \right . 
\ee
where $\bzt_{n, k}$ and $\bzc_{n, k}$ indicate the target and clutter measurements, respectively. The relationship between the measurement $\bzt_{n, k}$ 
 and the target state  is given by $h(\bx_k)$ and $\bv_{n,k}$ 
 is the measurement noise process. 
 
\begin{algorithm}[t]
\begin{algorithmic}[ ] 
\STATE Initialize target state: $\bx_0$ 
\FOR{$k =1:K$}
\STATE  Predict  $p\Big (\bx_k \mid  \bzzt_{k-1} \Big )$ using  \\
{\small{ $p\Big (\bx_k \mid  \bzzt_{k-1} \Big ) \eqq 
 \displaystyle  \int p(\bx_k \mid \bx_{k-1}) \; p\Big (\bx_{k-1} \mid \bzzt_{k-1}\Big ) d\bx_{k-1}$}}
%
\STATE Input measurements $ \{ \bz_{1, k}, \  \ldots, \  \bz_{N_k, k} \}$ \vskip 2pt
 \STATE Draw clutter prior $G^{(\text{c})}_k$ from $\text{DP}(\alpha_c, H_c)$  \vskip 2pt
\STATE Use clutter prior to draw 
	$\Theta_k \mid G^{(\text{c})}_k
	 \overset{\text{iid}}{\sim}{G^{(\text{c})}_k}$ \vskip 2pt
 \STATE Draw  target prior $G^{(\text{t})}_k$ from \\
 	\hspace{0.4cm}$G^{(\text{t})}_k \mid  \Theta_k  \sim 
	\text{DP} \Big(\alpha_t, H_t + \sum_{i=1}^{N_k} \delta_{\theta_{i, k}} \Big)$ \vskip 2pt
\STATE Use target prior to draw 
$W_k \mid G^{(\text{t})}_k, \Theta_k 
	\overset{\text{iid}}{\sim}  {G^{(\text{t})}_k}$ \vskip 2pt
\STATE  For  target measurements, draw
        $\bzt_{n, k}$, using Equation \eqref{zt}  \vskip 2pt
 \STATE For clutter measurements,  draw  
 $\bzc_{n, k}$, using Equation \eqref{zc}  \vskip 2pt
\STATE Cluster into target measurements $\bzzt_k$ with cardinality $M^{(\text{t})}_k$
\vskip 3pt
\STATE Cluster into clutter measurements $\bzzc_k$ with cardinality $M^{(\text{c})}_k$
\vskip 3pt
 \STATE  Compute  likelihood ratio  
 	$L\Big ( \bzzt_k; W_k, \bx_k  \Big )$ using Equation \eqref{likeli}
	\vskip 3pt
 \STATE  Compute and return the  posterior density using\\
	\hspace{2.1cm}$p\Big (\bzzt_k  \mid  \bx_k, W_k \Big )$\vskip 2pt
\STATE Update $p(\bx_k \mid  \bzzt_k)$
	using \\ \hspace{0.4cm}$p\Big (\bx_k \mid   \bzzt_k \Big )  \propto 
	p\Big (   \bzzt_k\mid W_k, \bx_k \Big ) \, p\Big ( \bx_k \mid \bzzt_{k-1}\Big )$ \vskip 2pt
\ENDFOR
\end{algorithmic}
\caption{METRIC-Bayes: Tracking in high clutter using jointly DPs prior.} 
\label{alg:DPtracking}
\end{algorithm}

Given the measurements $\bz_{n, k}$, $n\eqq 1, \ldots,N_k$,
we can obtain the target state by estimating  the posterior density 
$p\big (\bx_k \mid  \bz_{1, k}, \  \ldots, \  \bz_{N_k, k} \big )$.
However, if all available measurements are used, the presence of clutter deteriorates the tracking performance. In order to improve the performance, we propose the METRIC-Bayes approach, as 
described in Algorithm \ref{alg:DPtracking}.  In what follows, we discuss this framework in detail.

\subsection{Bayesian Nonparametric Prior}
\label{problem_prior}

 In this section, we discuss the framework summarized in Algorithm  \ref{alg:DPtracking} in details. Our framework models the marginal distributions of the joint prior as two conditionally independent DPs. 
Note that in this model, no prior information on the measurements is required.
Due to the joint measurement distribution factorization $ p\big(\bzt, \bzc\big) =  p\big(\bzt | \bzc\big)  p\big(\bzc\big) $, we set two conditionally independent Dirichlet processes on the parameters of $\bzc$ and $\bzt | \bzc$. Therefore, we may model the clutter prior independently as $G^{(\text{c})}_k \sim \text{DP}(\alpha_c, H_c)$  where DP is a Dirichlet process with scale parameter $\alpha_c$ and base distribution $H_c$. To model the target prior,  we first 
draw the sample set $\Theta_k \eqq \{ \theta_{1, k} \ldots, \theta_{N_k, k} \}$
from the clutter prior as 
$\Theta_k \mid  G^{(\text{c})}_k \overset{\text{iid}}{\sim}  G^{(\text{c})}_k$ and then given $\Theta_k$, 
the target prior is defined as  
\be
G^{(\text{t})}_k  \mid  \Theta_k  \sim  \text{DP}\Big (\alpha_t, 
 H_t + \sum_{n=1}^{N_k}   \delta_{\theta_{n, k}}  \Big ),
\label{true_meas_prior}
\ee
where $\delta(\theta_{n, k})$ is the Dirac delta  
 concentrated at $\theta_{n, k}$.
The  set $W_k \eqq \{ w_{1, k} \ldots, w_{N_k, k} \}$
is drawn from according to the Equation \ref{true_meas_prior} as  
$W_k \mid  G^{(\text{t})}_k   \overset{\text{iid}}{\sim} {G^{(\text{t})}_k  }$.
Using  sets $\Theta_k$ and $W_k$, we can then draw 
the target and clutter measurements, respectively,  as  
\be
& & \bzt_{n, k}  \mid  W_k, \Theta_k \  \sim \  F_t(\cdot \mid W_k), \; 
n=1, \ldots, N_k  \label{zt}  \\[.1in]
& & \bzc_{n, k}  \mid   W_k, \Theta_k \   \sim \  F_c(\cdot \mid \Theta_k ), \; 
n=1, \ldots, N_k.
\label{zc}
\ee
For some distributions $F_t$ and $F_c$. The generative model describing this process is summarized as 
\be
\label{generative}
\begin{split}
&\Theta_k \mid  G^{(\text{c})}_k \overset{\text{iid}}{\sim}  G^{(\text{c})}_k\\
&G^{(\text{t})}_k  \mid  \Theta_k  \sim  \text{DP}\Big (\alpha_t, 
 H_t + \sum_{n=1}^{N_k}   \delta_{\theta_{n, k}}  \Big )\\
 &\bzt_{n, k}  \mid  W_k, \Theta_k \  \sim \  F_t(\cdot \mid W_k), \; 
n=1, \ldots, N_k  \label{zt}  \\[.1in]
& \bzc_{n, k}  \mid   W_k, \Theta_k \   \sim \  F_c(\cdot \mid \Theta_k ), \; 
n=1, \ldots, N_k.
\end{split}
\ee

\subsection{Bayesian Tracker}
\label{problem_tracker}
Thus far, we modeled the joint distribution of measurements through the joint Dirichlet process. However, in order to get an accurate and efficient estimate, one needs to incorporate true measurements into the Bayesian tracker as follows. We cluster  the target measurements into  
set $\bzzt_k$ with cardinality $M^{(\text{t})}_k$ and the 
clutter measurements into  set $\bzzc_k$
 with  cardinality $M^{(\text{c})}_k$, where $M^{(\text{t})}_k + M^{(\text{c})}_k = N_k$.
 The likelihood ratio is thus computed as
\be 
L \Big ( \bzzt_k; W_k,  \bx_k  \Big ) =   
 \frac{\displaystyle \prod_{m=1}^{N_k}  
 p\Big ( \bz_{m, k}^{(t)} \mid  \bx_k; \rm{target\ present} \Big )} 
 {\displaystyle \prod_{m=1}^{N_k}
 p\Big ( \bz_{m, k}^{(c)} \mid \bx_k;  \rm{target\ absent} \Big )}.
 \label{likeli}
 \ee
Equation \eqref{likeli} is then incorporated into a Bayesian  tracker
to estimate the target state as 
\be
p\Big (\bx_k \mid   \bzzt_k \Big )  \propto p\Big (   \bzzt_k\mid W_k, \bx_k \Big ) \, p\Big ( \bx_k \mid \bzzt_{k-1}\Big ).
\ee

Joint Bayesian nonparametric prior allows us to effectively model the measurements at every time step without prior information. Any knowledge about the tracking scene or clutter may be included in the base measures $([\alpha_c,H_c],[\alpha_t,H_t])$ of the joint Dirichlet process prior. A Markov chain Monte Carlo method may be employed to sample from the posterior. This model can be easily generalized to track multiple objects by incorporating it into multiple object tracking techniques e.g., integrating dependent priors on the states introduced in \cite{Mor18, Mor19a} into the METRIC-Bayes framework.

\section{Expriments}
\label{sim}
\no In this section, we demonstrate the performance of the proposed method and compare our results to that of the traditional Bayesian filtering as well nearest-neighbor (NN) and probabilistic data association (PDA) filters. We show through simulations that accounting for clutters drastically improves the performance the tracker. Furthermore, we demonstrate that METRIC-Bayes framework which incorporates a joint nonparametric prior in Bayesian tracker outperforms existing methods such as nearest neighbor and PDA filters. In particular, our simulation results indicate that estimating the trajectory of a target is exceedingly benefited from incorporating METRIC-Bayes to learn the distribution of measurements that are originated from the target. 

We assume a linear Gaussian scenario so that we are able to compare our framework to the existing methods. The following linear Gaussian single-target model is used that was introduced in \cite{vo2008bayesian}. Assume that the state vector $\bx_{\ell, k}$ is given by  $\bx_{\ell, k} \eqq [x_{\ell, k} \  y_{\ell, k} \  \dot{x}_{\ell, k}\ \dot{y}_{\ell, k} ]^T$, $\ell\eqq 1, \ldots, N_k$, where $[x_{\ell, k} \  y_{\ell, k}]^T$, $[\dot{x}_{\ell, k} \ \dot{y}_{\ell, k}]^{T}$ are the 2-D position and 2-D velocity, respectively. We assume a linear Gaussian model with transition probability
\ben
p(\bx_k\mid\bx_{k-1}) = \mathcal{N}(\mu, \Sigma)
\een
where $\bmu \eqq A_{k}  \bx_{k-1}$ and covariance matrix  $\Sigma\eqq \text{diag}( [  \sigma^2 B_k B^T_k])$ where
\ben
A_{k}\!\!=\!\!
\begin{bmatrix}
1 &0&\Delta&0\\
0 & 1 & 0 & \Delta\\
0&0&1&0\\

0 & 0 & 0& 1
\end{bmatrix}\!\!,
B_k = \Delta\begin{bmatrix}
\frac{1}{2}  \\
\frac{1}{2}\\
1\\
1
\end{bmatrix}\!.
\een
with $\sigma \eqq 7$ m/s$^2$ and $\Delta = 1$. We assume that the probability of a target remaining at a scene during transitioning to be  $\text{P}_{\ell, k \mid k-1}\eqq 0.95$, for all $\ell$. 

We also assume a Gaussian likelihood for measurements originated from the target according to the likelihood 
\ben
p(\bz_{l,k}|\bx_k) = \mathcal{N}(H_k \bx_k), Q_k)
\een
where 
\ben
H_{k}\!\!=\!\!
\begin{bmatrix}
1 &0&0\\
0 & 1 & 0 
\end{bmatrix}\!\!,
Q_k = {\sigma^\prime}^2\begin{bmatrix}
1&0  \\
0 & 1
\end{bmatrix}\!
\een
where ${\sigma^\prime} = 10 m$. The observation region is the square $\mathcal{Z} = [-1000, 1000]\times[-1000,1000] $ (units are in m). We also consider Poisson clutter with parameter $\lambda_k = 5$. Given the problem setup, we put our framework to test in the following scenarios. In both cases, the results are obtained by 10,000 Monte Carlo (MC) runs on the same trajectory.

\subsection{METRIC-Bayes vs traditional Bayesian Filtering}

\begin{figure}[t]
\centering
 \resizebox{3.5in}{!}{\includegraphics{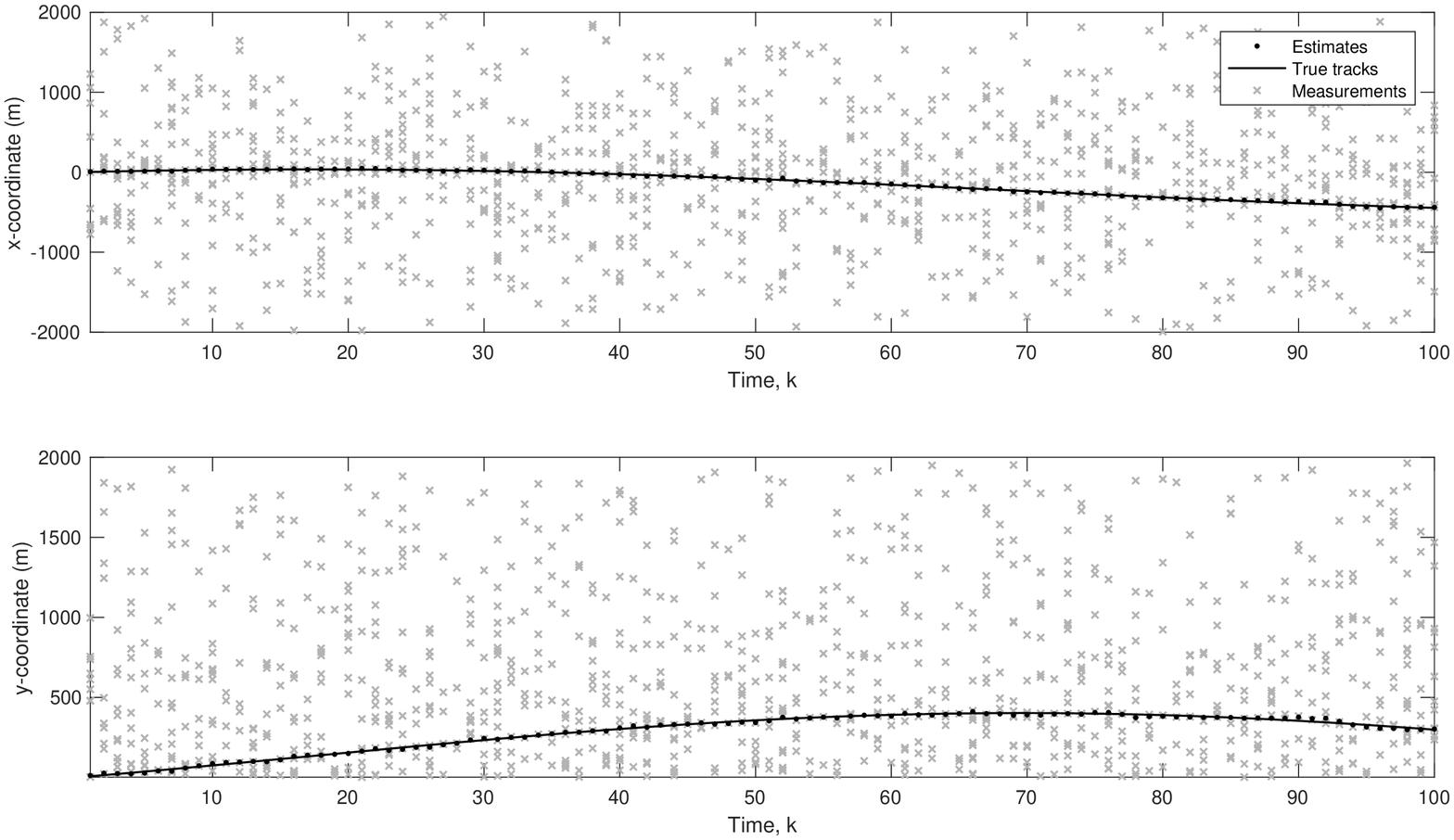}}
 \caption{True and estimate $x$-coordinate (top) and $y$-coordinate (bottom) using METRIC-Bayes.}
 \label{act_traj}
 \end{figure}

\begin{figure}[h]
\centering
 \resizebox{3.5in}{!}{\includegraphics{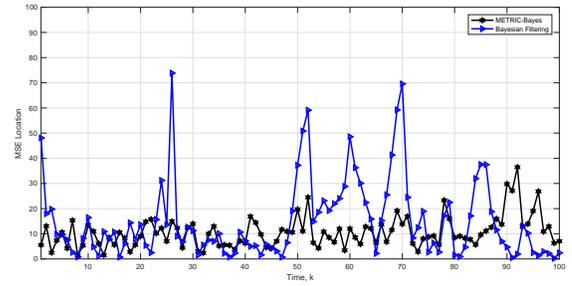}}
 \caption{Mean square error, METRIC-Bayes vs traditional Bayesian Filtering.}
 \label{MSE_TBT}
 \end{figure}

In this scenario, we compare the performance of our proposed framework to that of traditional Bayesian filtering. Traditional Bayesian filtering does not take into account the importance of clutters and consider them as one of the measurements. We compare the target location estimation mean-squared error (MSE) obtained using METRIC-Bayes vs Bayesian filter that uses all the measurements. A typical sample run of our proposed method is shown in Figure \ref{act_traj}, which suggests that the traditional filters tend to lose the track due to dealing with the clutters. Figure \ref{act_traj} also shows that we successfully estimate the trajectory of the target. 
Figure \ref{MSE_TBT} shows the mean square error (MSE) for the location of the target which signifies the importance of utilizing the actual measurements distribution to track the target. We computed the signal to the noise ration to be SCR = 5.9379. 

\subsection{METRIC-Bayes vs Nearest-Neighbor and PDA Filters}

\begin{figure}[h!]
\centering
 \resizebox{3.8in}{!}{\includegraphics{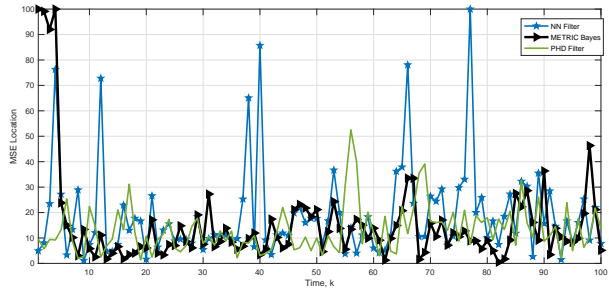}}
 \caption{Target location estimation mean-squared error (MSE) obtained using METRIC-Bayes vs NN and PDA filters for tracking a single object.}
 \label{General_comp}
 \end{figure}

In this section, we compare our proposed method to nearest-neighbor (NN) and probability data association (PDA) filters on the same set of data, which suggests that the NN and PDA filters tend to follow the pattern of extraneous target-generated measurements (clutter) and thus lead to poor performance. The mean square error (MSE) comparison in Figure \ref{General_comp} demonstrates that even in linear Gaussian case, the METRIC-Bayes method outperforms the NN and PDA filters. 
 
 It is worth mentioning we run 10'000 Monte Carlo runs for algorithms and algorithms are evaluated on the same dataset.

\newpage

\section{Conclusion}
\label{sec:Conc}

In this paper, we developed a nonparametric model for the measurements in the presence of high clutter. Our generative model is integrated into a Bayesian tracker to robustly track a target using time-varying measurements. We exploited the joint Dirichlet process prior to model the clutter and the measurements that are originated from the target. The output of our framework was then fed to a Bayesian tracker to estimate the trajectory of the target. METRIC-Bayes provided a robust and computationally efficient framework that can model any clutter distribution as well as efficiently estimating the target trajectory in high clutter with unknown number of clutters. We demonstrated through simulations that the amalgamation of the Bayesian nonparametric prior with a Bayesian tracker improves the tracking algorithm.

\bibliographystyle{IEEEtran}
\bibliography{refs_clutter}

\end{document}